\documentclass[
]{ceurart}

\usepackage{listings} 
\usepackage{mdframed}
\usepackage{placeins}
\usepackage{xcolor}
\newmdenv[
    backgroundcolor=white,
    linecolor=black,
    linewidth=1pt,
    topline=true,
    bottomline=true,
    rightline=true,
    leftline=true,
    roundcorner=5pt,
    skipabove=10pt,
    skipbelow=10pt,
    innertopmargin=10pt,
    innerbottommargin=10pt,
    frametitlefont=\bfseries,
    frametitlebackgroundcolor=black!10,
    frametitlerule=true,
    frametitlerulecolor=black,
    frametitleaboveskip=5pt,
    frametitlebelowskip=10pt
]{custombox}

\begin{document}

\copyrightyear{2024}
\copyrightclause{Copyright for this paper by its authors. Use permitted under Creative Commons License Attribution 4.0 International (CC BY 4.0).}

\conference{SCOLIA 2025, the First International Workshop on Scholarly Information Access, ECIR 2025, 10th April 2025, Lucca, Italy}

\title{Ask, Retrieve, Summarize: A Modular Pipeline for Scientific Literature Summarization}

\author[1]{Pierre Achkar}[orcid=0009-0007-0791-9078]
\author[2]{Tim Gollub}[orcid=0000-0003-1737-6517]
\author[3]{Martin Potthast}[orcid=0000-0003-2451-0665]

\address[1]{Leipzig University, Fraunhofer ISI Leipzig}
\address[2]{Bauhaus-Universität Weimar}
\address[3]{Kassel University, hessian.AI, ScaDS.AI}

\begin{abstract} The exponential growth of scientific publications has made it increasingly difficult for researchers to stay updated and synthesize knowledge effectively. This paper presents \textit{XSum}, a modular pipeline for multi-document summarization (MDS) in the scientific domain using Retrieval-Augmented Generation (RAG). The pipeline includes two core components: a question-generation module and an editor module. The question-generation module dynamically generates questions adapted to the input papers, ensuring the retrieval of relevant and accurate information. The editor module synthesizes the retrieved content into coherent and well-structured summaries that adhere to academic standards for proper citation. Evaluated on the \textit{SurveySum} dataset, XSum demonstrates strong performance, achieving considerable improvements in metrics such as CheckEval, G-Eval and Ref-F1 compared to existing approaches. This work provides a transparent, adaptable framework for scientific summarization with potential applications in a wide range of domains. Code available at https://github.com/webis-de/scolia25-xsum/tree/main
\end{abstract}

\begin{keywords}
  Multi-document Summarization (MDS) \sep Retrieval-Augmented Generation (RAG) \sep Scientific Literature Summarization
\end{keywords}

\maketitle

\section{Introduction}
The rapid growth of scientific literature has made it increasingly difficult for researchers to stay up-to-date with the latest developments. The number of papers published each month has increased exponentially since 1994, with fields such as artificial intelligence (AI) doubling their research output~ \cite{DBLP:journals/natmi/KrennBCEFGLLMSSTVXYK23}. While this growth reflects the progress of research communities, it also presents a serious challenge:~how can researchers stay informed and extract key insights from this volume of information? This overload of information makes it difficult to manually read, understand, and summarize the growing body of literature. This challenge becomes paramount in rapidly evolving fields such as AI, where researchers often need to synthesize knowledge from multiple sources in order to make progress. Summarizing research is not simply reading through papers but also identifying the most important information, connecting ideas from different sources, and presenting them in a clear and concise way. Automated summarization solutions are essential to help researchers save time and focus on the core information.\\
One promising approach to this challenge is Multi-Document Summarization (MDS), which combines information from multiple sources into clear and concise summaries. The concept itself is not new; for example, early work from 1999 proposed to use reference relationships between scientific papers to generate survey-style summaries \cite{DBLP:conf/ijcai/NanbaO99}. The approach identifies key fragments of cited papers, analyzes similarities and differences between them, and classifies citation contexts to support summarization. Over time, summarization methods have evolved from static approaches to deep learning models and later to pre-trained language models (PLMs)~\cite{DBLP:journals/corr/abs-2406-11289}. Currently, the field is dominated by Large Language Models (LLMs), which are pre-trained on massive datasets and capable of generating high-quality text.\\
Retrieval-Augmented Generation (RAG) builds on these advances by combining retrieval techniques with LLMs, enabling systems to find relevant information and synthesize it into accurate and coherent answers. A typical RAG pipeline processes a set of documents \( D = \{d_1, d_2, \dots, d_n\} \) by dividing them into smaller chunks, encoding them into dense vector embeddings with a pre-trained model, and storing them in a vector database for later retrieval. In the context of MDS, a search query to the vector database acts as a summarization guideline that can either be provided by the user or the MDS system. When a query is provided, the top-\( k \) most relevant chunks are retrieved based on similarity metrics and passed to an LLM, which generates responses grounded in the retrieved content.\\
Despite the recent advances, summarizing scientific literature remains an open research problem, requiring not only linguistic fluency and coherence, but also robust relevance and adherence to academic standards for citing literature correctly. To address these challenges, we present \textit{XSum}, a RAG pipeline designed for MDS in the scientific domain. XSum builds upon the typical RAG pipeline and introduces two new innovative components:~a question-generation module and an editor module. The question-generation module formulates questions on the basis of the input papers to be summarized, which are then passed to the RAG component. The editor module synthesizes the set of answers retrieved from the RAG component into coherent summaries, ensuring that the resulting output is comprehensive, reliable, and well-structured. The proposed pipeline is evaluated on the \textit{SurveySum} \cite{DBLP:journals/corr/abs-2408-16444} dataset, which is designed to test MDS methods in the scientific domain. The results show that XSum outperforms existing methods on metrics such as CheckEval \cite{DBLP:journals/corr/abs-2403-18771}, G-Eval \cite{DBLP:journals/corr/abs-2303-16634} and Ref-F1, demonstrating its ability to produce high-quality summaries. We consider the quality of a generated summary to be defined by its ability to comprehensively cover the essential content of the source documents, to maintain a coherent and fluent narrative, and to accurately reflect the original citations.

\section{Related Work}
The task of summarizing multiple scientific documents has evolved considerably over time. Early approaches, such as \textit{SciSumm}, introduced query-driven summarization by clustering relevant text segments from co-cited papers to generate contextualized summaries \cite{agarwal-etal-2011-scisumm}. These methods leveraged citation relationships but struggled with complex content relationships across documents.

Later developments introduced neural network-based architectures for MDS. For instance, \textit{HiMAP} and \textit{HierSumm} utilized hierarchical models and passage ranking techniques to enhance content selection and fusion, resulting in more coherent and contextually relevant summaries \cite{fabbri-etal-2019-multi, liu-lapata-2019-hierarchical}. These methods marked a shift from purely extractive approaches to more integrative models capable of generating fluent summaries.

The integration of extraction and abstraction further refined summarization methods. Shinde et al. proposed a hybrid pipeline that combines BERT-based extractive models with \textit{BigBird-PEGASUS} for abstractive summarization, achieving robust performance in the biomedical domain \cite{shinde-etal-2022-extractive}. Similarly, \textit{KGSum} introduced knowledge graph-based encoding to model document content and relationships, employing a two-stage decoding strategy to produce focused and cohesive summaries~\cite{DBLP:conf/coling/WangLPHLT022}.

The field has taken a major step forward with the emergence of retrieval augmented generation (RAG) pipelines. \textit{OpenScholar}\footnote{\url{https://openscilm.allen.ai/}}, for example, demonstrated a novel approach by integrating a specialized datastore of 45 million papers with iterative retrieval and feedback loops, enabling precise, citation-backed responses, highlighting growing interest in retrieval augmented systems \cite{DBLP:journals/corr/abs-2411-14199}. Another approach to MDS using retrieval is proposed through the \textit{SurveySum} framework, which introduces two pipelines, Pipeline~1 and Pipeline~2, both integrating retrieval-based selection with LLM-based summarization \cite{DBLP:journals/corr/abs-2408-16444}. These pipelines are evaluated on the \textit{SurveySum} dataset, a benchmark specifically designed for MDS in scientific literature, which consists of survey sections paired with their cited papers. This is the same dataset used in this work, and a more detailed discussion of its structure will be provided in the Evaluation section. 

\begin{figure}[ht]
    \centering
    \includegraphics[width=0.8\textwidth]{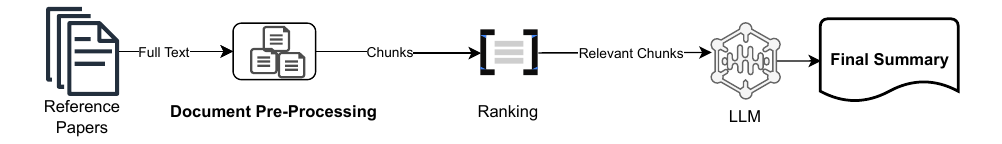}
    \caption{Overview of Pipeline 1: The system segments full-text papers into overlapping chunks, ranks them using \textit{monoT5-3B} based on the section title, and selects the top-ranked chunks for LLM-based summarization.}
    \label{fig:pipeline1}
\end{figure}

\begin{figure}[ht]
    \centering
    \includegraphics[width=0.8\textwidth]{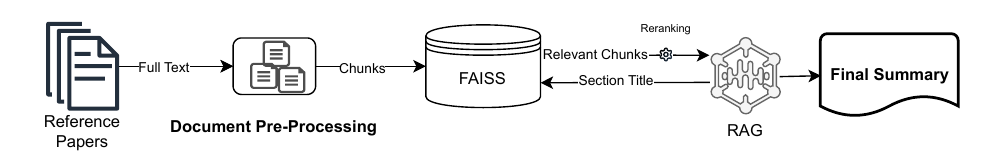}
    \caption{Overview of Pipeline 2: Instead of ranking with a neural model, this pipeline encodes chunks as dense embeddings using \textit{SPECTER2}, stores them in a \textit{FAISS} vector database, retrieves them based on section title queries, and applies reranking before LLM-based summarization.}
    \label{fig:pipeline2}
\end{figure}

Pipeline 1 uses a neural ranking approach where full-text papers are segmented into overlapping chunks during pre-processing. These chunks are ranked by \textit{monoT5-3B}, which assigns relevance scores based on the title of the target survey section. The highest-ranked chunks are then passed to an LLM, such as \textit{GPT-4}, to generate the final summary. Pipeline 2, on the other hand, relies on embedding-based retrieval, where text chunks are represented as dense embeddings using \textit{SPECTER2}\footnote{\url{https://huggingface.co/allenai/specter2}} and stored in a \textit{FAISS} vector database. The section title (e.g. Data Generation via PLM:Explaining Models’ Decisions) is used as a query to retrieve relevant chunks at inference time. Unlike Pipeline 1, which directly selects the top-ranked chunks for summarization, Pipeline 2 includes a re-ranking step where an LLM evaluates and ranks the retrieved content before summarization. The resulting chunks are then summarized into a cohesive section. Figures~\ref{fig:pipeline1} and~\ref{fig:pipeline2} illustrate the structures of these pipelines.  

While these pipelines achieve acceptable performance, they rely on static retrieval using section titles as queries, which can limit adaptability to different summarization contexts. Among them, Pipeline 2 is more comparable to our approach \textit{XSum}, as it utilizes embedding-based retrieval rather than direct ranking. However, \textit{XSum} addresses key limitations by introducing a question generation module that dynamically formulates structured questions based on the title and abstract of the input papers, which serve as queries during retrieval, thereby improving retrieval relevance. Additionally, it features an editor module that synthesizes retrieved content into a coherent, citation-rich summary, ensuring better fluency, accuracy, and adherence to academic writing standards. The complete pipeline and the functionality of these components will be explained in detail in the Methodology section.

\noindent
Beyond \textit{SurveySum}, several other datasets have been developed for MDS, particularly in the biomedical domain. Datasets such as \textit{Cochrane-auto} and \textit{MS2} focus on summarizing clinical trials and systematic reviews, providing benchmarks for evaluating summarization methods in evidence-based medicine \cite{bakker-kamps-2024-cochrane, DBLP:journals/corr/abs-2104-06486}. Another relevant dataset is \textit{Multi-XScience}, which was initially considered for evaluating the proposed approach, as it focuses on synthesizing related work sections from abstracts and cited references \cite{DBLP:conf/emnlp/LuDC20}. However, a preliminary analysis revealed missing values in the reference papers used to generate the related work sections, raising concerns about its completeness for reliable benchmarking. Furthermore, while related work sections can be considered multi-document summaries, they are often shaped by the comparative and argumentative nature of the paper's contributions rather than being purely extractive or abstractive. Given these considerations, the \textit{SurveySum} dataset appeared to be a more appropriate choice for evaluating our approach, as it explicitly focuses on summarizing multiple scientific papers into structured survey sections. To our knowledge, no other experimental work has been conducted on \textit{SurveySum} beyond the evaluations presented by its authors so far.

\section{Methodology}  
This section introduces the \textit{XSum} pipeline, a modular approach to summarizing scientific literature into coherent and traceable outputs. The initial idea for building this pipeline was inspired by the interview paradigm, where an interviewer interacts with a domain expert. In this analogy, the interviewer prepares a structured set of questions based on the expert's domain knowledge, conducts the interview in which the expert answers these questions, and finally an editor compiles the conversation into a well-structured summary. This concept motivated the design of \textit{XSum} and led to the introduction of two key modules: a question generation module, which formulates structured questions to guide the retrieval process, and an editor module, which synthesizes the retrieved answers into a coherent and citation-rich summary. Each module plays an important role in ensuring that the summaries generated are both relevant and well-structured, as described in the following subsections.

\subsection{Overview of the Pipeline}
\begin{figure}[ht]
\centering
\includegraphics[width=\textwidth]{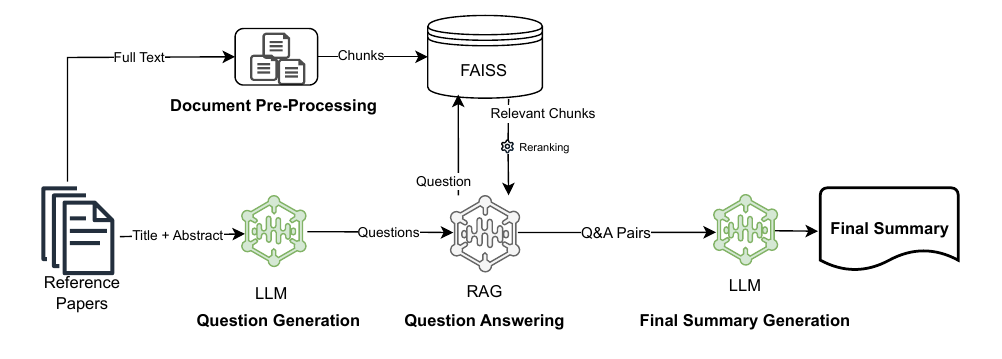}
\caption{Overview of the XSum Pipeline. The pipeline processes reference papers into summaries through modular steps. Document Pre-Processing segments papers into chunks, encodes them as embeddings, and stores them in a FAISS database. Question Generation uses an LLM to generate questions from titles and abstracts. In Question Answering, a RAG framework retrieves relevant chunks and generates answers with an LLM. Finally, the Editor Module (Final Summary Generation) synthesizes the answers into a coherent, citation-rich summary.}
\label{fig:pipeline}
\end{figure}

The proposed pipeline for MDS, \textit{XSum}, illustrated in Figure~\ref{fig:pipeline}, transforms input reference papers into a coherent summary through a sequence of modular steps. It begins by using the titles and abstracts of the reference papers to generate broad and general questions using an LLM. These questions, designed to reflect the main themes and contributions of the papers, are stored for later use. The full texts of the reference papers are then processed by dividing them into manageable chunks, which are embedded in dense vector representations and stored in a vector database. This pre-processing ensures efficient retrieval of relevant content in subsequent stages. The stored questions are then used to query the database and retrieve the most relevant chunks. The retrieval process involves an initial similarity-based ranking of the chunks, followed by a re-ranking step to refine their relevance. The final set of retrieved chunks is paired with the corresponding questions. These question-chunk pairs are then passed to an LLM, which generates concise answers based on the retrieved content. If the context is insufficient, the LLM will refrain from generating an answer, ensuring accuracy and credibility. Finally, the set of question-answer pairs is passed to the editor module, which synthesizes them into a comprehensive and well-structured summary. The editor ensures coherence, logical flow, and adherence to academic standards while incorporating citations to maintain traceability.

\subsection{Question Generation Module}
This module is essential for aligning the retrieval and summarization stages with the specific content of the reference documents. By leveraging the generative capabilities of LLMs, it ensures that the pipeline is driven by structured, contextually relevant queries. The approach draws on insights from methods such as \textit{HyDe} \cite{DBLP:journals/corr/abs-2212-10496}, \textit{HyQE} \cite{DBLP:journals/corr/abs-2410-15262} and \textit{reverse HyDe} \cite{DBLP:journals/corr/abs-2312-10997}, all of which use generative techniques to improve retrieval relevance. \textit{HyQE} (Hypothetical Document Embeddings) involves generating hypothetical content based on a query, encoding this content into embeddings, and then using these embeddings to improve retrieval accuracy. Both \textit{HyQE} (Hypothetical Query Embeddings) and \textit{reverse HyDe} follow a similar strategy, but focus on generating hypothetical questions or queries that match the content of a document. These hypothetical questions bridge the semantic gap between queries and retrievable content, improving the ranking of relevant results \cite{DBLP:journals/corr/abs-2410-15262}. 

In our pipeline, the title and abstract of each reference paper serve as input to a pre-trained LLM, which generates $k=5$ broad and semantically rich questions encapsulating the core themes and contributions of the paper. These questions are stored as structured queries for subsequent stages. The generated questions serve two primary functions: first, they refine the retrieval process by ensuring that only the most contextually relevant content is retrieved; second, they provide a structured framework to guide the subsequent synthesis and summarization phases. For illustration, examples of such generated questions can be found in Appendix \ref{appendix:questions_examples}.

\subsection{Document Pre-processing}
The pre-processing phase ensures that the reference papers are prepared for efficient retrieval and summarization by arranging them in a format suitable for downstream tasks. This phase consists of three main steps:

\begin{itemize}
    \item \textbf{Chunking Documents:} The full texts of the reference papers are divided into interconnected chunks of 150 tokens each, with an overlap of 20 tokens. This overlap preserves contextual continuity between successive chunks, while respecting sentence boundaries ensures that the division does not disrupt the semantic flow of the text. We determined this configuration by experimentation, after trying different setups, finding that it provided the best balance between contextual preservation and computational efficiency.
    \item \textbf{Embedding Generation:} Each chunk is encoded into dense vector representations using the \textit{SPECTER2} model, which is specifically designed to capture the semantic relationships and contextual meanings in academic texts. 
    \item \textbf{Vector Database Indexing:} Chunks are indexed in \textit{FAISS}, a high-speed similarity search database, for efficient retrieval.
\end{itemize}

\subsection{Question Answering Module}
In this module, the focus is on integrating retrieval and synthesis to generate concise, contextually relevant answers to the questions formulated in the previous stage. By combining robust retrieval techniques with an LLM in a RAG framework, this module ensures that the pipeline produces high-quality output that is grounded in the source material.

Questions are embedded into dense vector representations using the same \textit{SPECTER2} model used in the document pre-processing phase. The retrieval process proceeds in two stages:
\begin{enumerate}
    \item \textbf{Initial Retrieval:} Using cosine similarity, the top 100 chunks most relevant to each question are retrieved from the \textit{FAISS} vector database, serving as an initial filtering step.
    \item \textbf{Reranking:} The retrieved chunks are re-ranked using the \textit{ColBERT2} model \cite{DBLP:journals/corr/abs-2112-01488}, which evaluates token-level interactions between the question and the chunks. This refinement step ensures that the 20 most relevant chunks are selected.
\end{enumerate}

The final set of 20 chunks is presented to a pre-trained LLM along with the corresponding question. The LLM synthesizes a coherent and accurate response based solely on the retrieved context. If the retrieved chunks do not provide sufficient information, the LLM is instructed not to generate an answer, minimizing unsupported or speculative output.

To ensure credibility and traceability, the LLM includes valid citations from the retrieved chunks in its responses. By grounding the answers in the source material, this module adheres to academic standards and facilitates the verification of the generated content.

\subsection{Final Summary Generation (Editor Module)}
The Editor Module synthesizes the answers generated in the previous step into a cohesive and comprehensive summary, aggregating all question-answer pairs into a unified narrative that reflects the overarching themes and contributions of the papers. A pre-trained LLM is used as the editor to generate the final summary. The model is prompted to write an extensive, coherent summary that seamlessly integrates the individual answers while maintaining a logical flow. The Editor LLM ensures the summary adheres to academic standards. It incorporates citations into the final summary, ensuring that all statements are properly grounded in the retrieved source material. 
The prompt used in this module is as follows:
\begin{custombox}[frametitle=Editor Module Prompt]
\begin{lstlisting}
### CONTEXT ###
You are writing the final script of an interview with an expert on the topic '{topic}'. 
The final script should summarize the key insights and findings from the questions and answers provided. 
Keep the target audience in mind, which includes researchers, students, and professionals in the field.

### QUESTIONS AND ANSWERS ###
{questions_and_answers}

### INSTRUCTIONS ###
Include the most relevant and important points discussed.
Be aware of plagiarism, i.e., you should not copy the text, but use them as inspiration.
Avoid using markdown formatting in the text.
Avoid splitting into subsections, or creating an introduction and conclusion for it.
Avoid introducing new information and focus on summarizing the existing content.
Always include the citations (e.g., [BIBREF14], [BIBREF16]) mentioned in the answers in the final section.
\end{lstlisting}
\end{custombox}

\section{Evaluation}
The proposed pipeline is evaluated using a domain-specific dataset for MDS. This section details the dataset, metrics, implementation, results, examples, and discussion, providing a comprehensive analysis of the pipeline's performance.

\subsection{Dataset}
The evaluation of the proposed pipeline is conducted on the \textit{SurveySum}\footnote{\url{https://github.com/unicamp-dl/surveysum}} dataset, a domain-specific resource designed for MDS tasks in scientific literature. This dataset includes 79 survey sections across fields such as AI, natural language processing (NLP), and machine learning (ML). Each section is paired with the full-text content of its cited papers, with an average of 7.38 papers cited per section. The dataset is explicitly designed to test MDS models on the synthesis of content from multiple sources, making it particularly suited for assessing the proposed pipeline.

\subsection{Metrics}
The evaluation employs a mix of traditional and LLM-based metrics to assess the quality of summaries in terms of content coverage, coherence, and citation alignment:\\

\textbf{ROUGE} (Recall-Oriented Understudy for Gisting Evaluation) \cite{lin-2004-rouge} measures the overlap between the generated summaries and the reference text. It calculates n-gram overlap, word sequence matching, and the longest common subsequences. ROUGE-1, ROUGE-2, and ROUGE-L are used in this study to capture unigrams, bigrams, and sentence-level matches, respectively.\\ 
\textbf{BERTScore} \cite{DBLP:journals/corr/abs-1904-09675} evaluates semantic similarity between the generated and reference summaries using contextual embeddings from PLMs like BERT.\\ 
\textbf{Reference F1 Score} (Ref-F1) measures how accurately the citations in the generated summaries align with those in the ground truth. It computes precision (proportion of correctly included references) and recall (proportion of ground-truth references captured in the generated summary), and combines them into an F1 score. This metric is essential in scientific summarization, where attribution and citation accuracy are critical.\\
\textbf{G-Eval}\footnote{\url{https://github.com/nlpyang/geval}} \cite{DBLP:journals/corr/abs-2303-16634} is a framework for evaluating the output of natural language generation (NLG) using LLMs, providing reference-free assessments based on criteria such as coherence, coverage, fluency, and relevance. It uses Chain-of-Thought (CoT) reasoning to systematically generate detailed evaluation steps, ensuring consistency and robustness in scoring. Scores are assigned on a fixed scale (e.g. 1 to 5) and refined using token probabilities, enabling granular, continuous analyses that capture the nuances between outputs. By bypassing the need for reference outputs, it is particularly effective for tasks where predefined references are unavailable or impractical, such as creative or open-ended text generation. Experiments show that G-Eval achieves a stronger correlation with human judgments than traditional metrics such as ROUGE, as well as neural evaluators, on benchmarks such as \textit{SummEval} \cite{DBLP:journals/tacl/FabbriKMXSR21} for summarization.\\
\textbf{CheckEval}\footnote{\url{https://github.com/jayralencar/check-eval}} \cite{DBLP:journals/corr/abs-2403-18771} is a robust evaluation framework that uses LLMs to evaluate generated text using a structured checklist-based approach. It supports two evaluation modes: reference-based, which compares the generated text to reference summaries, and criteria-driven, which evaluates the text against predefined dimensions such as coherence, fluency, and coverage. By breaking down evaluation criteria into detailed sub-aspects, framed as Boolean (yes/no) questions, CheckEval simplifies the evaluation process and increases its reliability and interpretability. The framework operates in three stages: aspect selection, where key evaluation dimensions are identified; checklist generation, where detailed questions are created and refined; and checklist-based evaluation, where LLMs respond to the questions, with the final score calculated as the proportion of positive responses. Validated against the \textit{SummEval} benchmark, CheckEval demonstrates high correlation with human judgment and strong inter-annotator agreement.\\

ROUGE and BERTScore evaluations report recall scores to emphasize content coverage, while G-Eval and CheckEval focus on the coverage criterion, consistent with SurveySum's methodology for assessing core content representation.

\subsection{Implementation Details}
This section outlines the tools, models, and frameworks utilized in our development process:

\begin{itemize}
    \item \textbf{Development Environment:} The pipeline was implemented in Python, utilizing \texttt{sentence-transformers} for embedding generation, \texttt{nltk} for text processing, and \textit{FAISS} for efficient vector-based retrieval. All experiments were conducted on a Tesla V100-PCIE-32GB GPU, enabling efficient embedding generation, chunk retrieval, and summarization tasks.  
    
    \item \textbf{Pre-trained LLMs:} For text generation, we employed \texttt{gpt4o-mini\_15-2-2024-preview}, while \texttt{Phi-3-small-8k-instruct} was utilized for evaluation. Both models were configured with a temperature of 0.3 and a top-p of 0.95 to ensure controlled and consistent outputs. This choice aligns with the position that the same model should not be used for both generation and evaluation to mitigate potential bias. Research has highlighted that using identical or equally powerful models for both tasks can lead to skewed results, as LLMs like GPT-4 tend to favor their own outputs due to egocentric biases \cite{DBLP:conf/emnlp/LiXSXGLTM24}. Notably, the \textit{SurveySum} paper does not specify whether identical models were used in their study for both tasks.  
\end{itemize}

\subsection{Results}
Since the original \textit{SurveySum} paper did not report ROUGE and BERTScore metrics for the benchmark pipelines, we calculated these values. G-Eval and CheckEval were also computed using our implementation to ensure consistency, employing the \texttt{Phi-3-small-8k-instruct} model as the evaluator. This approach guarantees a fair comparison across all pipelines, enabling a comprehensive assessment of XSum’s performance relative to the benchmarks. Any differences in the results (particularly for G-Eval and CheckEval) can be attributed to differences in the evaluation settings and model configurations compared to the original SurveySum experiments. For clarity, \textit{Pipeline\_1} and \textit{Pipeline\_2} are the two pipelines that performed best in the \textit{SurveySum} experiments.

The results, summarized in Table~\ref{tab:quantitative_results}, highlight XSum's consistent outperformance of the benchmark pipelines across all metrics. It achieves ROUGE-1 (0.51) and ROUGE-L (0.24), reflecting its ability to effectively capture unigrams and sentence-level structures. Its BERTScore (0.62) highlights its strong semantic alignment with reference summaries, reflecting its capability to retain content integrity through paraphrasing and semantic rephrasing. Furthermore, XSum attains the highest Ref-F1 (0.76), G-Eval (4.2), and CheckEval (0.97) scores, emphasizing its superiority in generating coherent, relevant, and high-quality summaries.

\begin{table*}
  \caption{Comprehensive performance comparison of the evaluated pipelines based on traditional metrics (ROUGE, BERTScore, Ref-F1) and LLM-based metrics (G-Eval, CheckEval).}
  \label{tab:quantitative_results}
  \begin{tabular}{lccccccc}
    \toprule
    \textbf{Pipeline} & \textbf{ROUGE-1} & \textbf{ROUGE-2} & \textbf{ROUGE-L} & \textbf{BERTScore} & \textbf{Ref-F1} & \textbf{G-Eval} & \textbf{CheckEval} \\
    \midrule
    Pipeline\_1 & 0.42 & 0.08 & 0.19 & 0.57 & 0.64 & 3.1 & 0.61 \\
    Pipeline\_2 & 0.49 & \textbf{0.10} & 0.23 & 0.59 & 0.72 & 4.0 & 0.76 \\
    XSum & \textbf{0.51} & \textbf{0.10} & \textbf{0.24} & \textbf{0.62} & \textbf{0.76} & \textbf{4.2} & \textbf{0.97} \\
    \bottomrule
  \end{tabular}
\end{table*}

To further demonstrate the performance of XSum, we present two examples of summaries generated by it, along with their corresponding ground truth (original section text) and evaluation scores. These examples have been selected based on their performance across the evaluation metrics, one representing the highest average score across all metrics and the other representing the lowest average score. Due to their length, the full examples are provided in Appendix~\ref{appendix:evaluation_examples}.

\subsection{Discussion}
The strong performance of \textit{XSum} is largely driven by its two key features: the question-generation module and the editor module. By dynamically generating queries about the document content, the retrieval module ensures relevant and contextual results, addressing the limitations of static query approaches such as using section titles, as in \textit{Pipeline\_2}. Additionally, the use of \textit{ColBERT} as a reranker may contribute to better chunk retrieval by prioritizing the most relevant and informative sections during the ranking process. The editor module further enhances the pipeline by synthesizing retrieved information into coherent summaries with proper citations, resulting in outputs that adhere to academic standards.

The quality of the content generated by \textit{XSum} highlights its ability to synthesize multiple sources into a structured and coherent summary. The selected examples illustrate both the strengths and limitations of the approach. The high-scoring example (Example 1) closely follows the human-written text, effectively capturing key technical details while maintaining logical flow and factual consistency. This suggests that XSum can generate summaries that are both informative and well-structured and in line with academic standards. However, a notable difference remains in the style and clarity of the summaries. Human-written sections tend to be more compact and nuanced, often presenting a comparative perspective that sets different contributions in relation to each other. In contrast, XSum summaries tend to be verbose, often providing extended explanations and additional contextual information beyond what is strictly necessary for summarizing. This is particularly evident in the low-scoring example (Example 2), where the generated text, while factually accurate, lacks the same level of selectivity as the human-written version, including an unnecessary degree of background detail rather than focusing solely on comparative insights. This contrast highlights a key challenge in scientific summarization. Although retrieval-driven methods such as XSum excel at aggregating and structuring information, they do not yet fully replicate the complex synthesis and prioritization that domain experts perform when writing a summary of multiple related papers. Nevertheless, XSum still produces highly structured and factually based summaries, demonstrating that automated MDS can be a valuable tool for scientific literature synthesis, particularly in assisting researchers with information overload.

Despite its strengths, the low ROUGE-2 scores across all pipelines highlight a common problem in abstractive summarization: achieving bigram overlap with reference summaries. RAG-based pipelines, including XSum, prioritize semantic richness and coherence over strict lexical matching, which reduces alignment with reference summaries. However, ROUGE-1 and ROUGE-L scores show moderate alignment, reflecting the ability to capture essential unigrams and sentence-level structures. BERTScore, which assesses semantic similarity, achieves satisfactory results, highlighting the ability of such pipelines to capture the essence of content through paraphrasing and semantic rephrasing, even when lexical overlap is limited. It is important to note that while the ROUGE metrics provide valuable insights into lexical overlap and content coverage, they are inherently limited in assessing the nuances of abstractive summarisation. This limitation is addressed by incorporating a set of metrics - BERTScore, G-Eval and CheckEval - that more effectively capture semantic similarity, coherence and overall quality.

In addition to traditional metrics, frameworks like G-Eval and CheckEval provide refined assessments of summary quality by leveraging LLMs to evaluate coherence, relevance, and coverage. These metrics excel at capturing semantic and structural attributes that conventional metrics often overlook, making them particularly effective for evaluating abstractive summaries. However, their dependence on specific LLMs introduces challenges of consistency and reproducibility, as evaluation outcomes may vary with different model configurations. This highlights the need for standardization in LLM-driven evaluation practices.

Finally, XSum's modular design offers substantial flexibility in adapting to different summarization tasks. The question-generation module can be customized to generate domain-specific or task-specific questions to improve relevance in different contexts. Similarly, the editor module allows customization of tone, style, and abstraction levels, enabling outputs to be tailored for different audiences, from academic researchers to professional practitioners. This adaptability ensures the pipeline’s scalability and applicability to a wide range of domains, addressing the growing demand for efficient MDS in complex settings.

\section{Conclusion and Future Work}
This work addresses the challenges of MDS in the scientific domain by introducing a modular RAG-based pipeline featuring two key enhancements: a question-generation module and an editor module. These components enable the pipeline to synthesize information from multiple scientific papers into cohesive, well-structured summaries. Experimental evaluations on the \textit{SurveySum} dataset demonstrate considerable improvements in metrics such as CheckEval, G-Eval, and Ref-F1 compared to existing approaches. By providing detailed guidance on the design and implementation of RAG-based pipelines, this work contributes to making these systems more transparent, reproducible, and adaptable for diverse summarization tasks.

While the current pipeline achieves strong performance, several opportunities for future improvements remain. A key direction is evaluating \textit{XSum} against other MDS pipelines to enable a more in-depth comparison of effectiveness and retrieval quality. Such comparisons would provide insights into the relative strengths and limitations of different summarization strategies. Additionally, conducting an ablation study would allow for a deeper understanding of the impact of each component in the pipeline, particularly the question-generation module and the editor module, to assess their individual contributions to overall performance. Optimizing data ingestion pipelines, which are often a bottleneck in large-scale industrial applications (as emphasized in systems like \textit{ColPali} \cite{DBLP:journals/corr/abs-2407-01449}), could further enhance scalability and efficiency. Moreover, integrating vision-language models to process visually rich documents, including text, tables, and figures, offers a promising direction for improving retrieval accuracy and extending the system's capabilities to more complex scientific datasets.

\section{Limitations}
Despite its contributions, this work has several limitations that warrant further investigation:

\begin{itemize} 
    \item \textbf{Scalability and Real-World Deployment:} While the proposed pipeline demonstrates strong performance in controlled environments, this work does not address the challenges of scaling the pipeline for real-world applications. Issues such as handling extremely large datasets, ensuring low latency, and optimizing cost-effective deployment for different organizational needs remain unaddressed and require further research.
    \item \textbf{Qualitative Analysis:} While quantitative evaluations on metrics like CheckEval and G-Eval demonstrate strong performance, this study lacks a comprehensive qualitative analysis of the generated summaries. 
    \item \textbf{Document Retrieval Scope:} The pipeline assumes a predefined set of input papers for summarization and does not address the challenge of identifying or retrieving relevant documents for a specific topic. This limitation highlights the need for further research into integrating robust document retrieval mechanisms with summarization workflows to enhance the pipeline’s applicability.
\end{itemize}

\newpage
\bibliography{references} 

\newpage
\appendix
\section{Appendix}
\subsection{Examples of Generate Questions}
\label{appendix:questions_examples}
Below are examples of generated questions produced by our question-generation module. 

\begin{custombox}[frametitle=Generated Questions Example 1] 

\textbf{Paper Title:} Automatic melody harmonization with triad chords: A comparative study\\

\noindent
\textbf{Paper Abstract:} Several prior works have proposed various methods for the task of automatic melody harmonization, in which a model aims to generate a sequence of chords to serve as the harmonic accompaniment of a given multiple-bar melody sequence. In this paper, we present a comparative study evaluating and comparing the performance of a set of canonical approaches to this task, including a template matching based model, a hidden Markov based model, a genetic algorithm based model, and two deep learning based models. The evaluation is conducted on a dataset of 9,226 melody/chord pairs we newly collect for this study, considering up to 48 triad chords, using a standardized training/test split. We report the result of an objective evaluation using six different metrics and a subjective study with 202 participants.\\

\noindent
\textbf{Generated Questions:} \begin{itemize} \item What are the key differences in performance among the various models evaluated for automatic melody harmonization? \item How does the dataset of 9,226 melody/chord pairs contribute to the robustness of the study's findings? \item What specific metrics were used for the objective evaluation of the models, and how do they compare in terms of effectiveness? \item What insights were gained from the subjective study involving 202 participants regarding the perceived quality of the harmonizations? \item What future directions for research in automatic melody harmonization does this study suggest based on its findings? \end{itemize} \end{custombox}

\FloatBarrier

\begin{custombox}[frametitle=Generated Questions Example 2] 
\textbf{Paper Title:} Virtuosonet: A hierarchical rnn-based system for modeling expressive piano performance\\

\noindent
\textbf{Paper Abstract:} In this paper, we present our application of deep neural networks to modeling piano performance, which imitates the expressive control of tempo, dynamics, articulations, and pedaling from pianists. Our model consists of recurrent neural networks with hierarchical attention and a conditional variational autoencoder. The model takes a sequence of note-level score features extracted from MusicXML as input and predicts piano performance features of the corresponding notes. To render musical expressions consistently over long-term sections, we first predict tempo and dynamics at the measure level and, based on the result, refine them at the note level. The evaluation through listening tests shows that our model achieves a more human-like expressiveness compared to previous models. We also share the dataset used for the experiment.\\
\\

\noindent
\textbf{Generated Questions:} \begin{itemize} \item What are the key components of the hierarchical RNN architecture used in Virtuosonet for modeling expressive piano performance? \item How does the model differentiate between measure-level and note-level predictions in terms of tempo and dynamics? \item What specific features from MusicXML are utilized as input for the model, and how do they contribute to the performance output? \item In what ways does the evaluation through listening tests demonstrate the model's superiority in expressiveness compared to previous approaches? \item What insights can be drawn from the dataset shared in the paper regarding the training and evaluation of deep music generation models? \end{itemize} \end{custombox}

\subsection{Evaluation Examples}
\label{appendix:evaluation_examples}

\begin{custombox}[frametitle=Evaluation Example 1]
\textbf{Survey Title:} AMMUS:A Survey of Transformer-based Pretrained Models in Natural Language Processing

\noindent
\textbf{Survey Section:} Taxonomy:Extensions:Long-Sequence T-PTLMs

\noindent
\textbf{Evaluation Scores:} ROUGE-1: 0.61, ROUGE-2: 0.19, ROUGE-L: 0.32, BERTScore: 0.69, Ref-F1: 1.00, G-Eval: 4.8, CheckEval: 1.00\\

\noindent
\textbf{Ground Truth (Section Text):}\\
The self-attention attention module in transformers updates the representation of each input token by attending to all tokens in the input sequence. The quadratic time complexity of the self-attention module limits the application of T-PTLMs to long input sequences. To overcome this drawback, self-attention variants like sparse self-attention and linearized self-attention are proposed to reduce its complexity and hence extend T-PTLMs to long input sequences also BIBREF205 . Some of the popular T-PTLMs based on a) sparse self-attention are Longformer BIBREF206 , ETC BIBREF207 , BigBird BIBREF208 and Reformer BIBREF209 and b) linearized self-attention are Performer BIBREF210 . Sparse self-attention reduces the complexity by including sparsity bias which reduces the number of query-key pairs that each query attends to. In linearized self-attention, reduced complexity is achieved by disentangling the attention with kernel feature maps and then computing the attention in reverse order.\\

\noindent
\textbf{Generated Content (XSum Output):} \\
The exploration of Transformer-based pretrained models, particularly in the context of long-sequence processing, has revealed significant advancements and innovations across various architectures. Key modifications to the original Transformer model have emerged, including Adaptive Computation Time (ACT), hierarchical structures, and dynamic halting mechanisms, all aimed at enhancing computational efficiency and representation capabilities for complex inputs [BIBREF205]. The introduction of models like Longformer, with its linear scaling attention mechanism and local-global attention integration, demonstrates a marked improvement in handling lengthy documents compared to traditional self-attention approaches, outperforming models such as RoBERTa in tasks requiring extensive context [BIBREF206].  The Extended Transformer Construction (ETC) architecture further addresses the challenges of scaling input length and encoding structured data through its global-local attention mechanism and the use of relative position encodings, which significantly enhance the model's performance on tasks like HotpotQA and WikiHop [BIBREF207].Similarly, BigBird's sparse attention mechanism allows for efficient processing of long sequences while maintaining the representational power of traditional Transformers, achieving state-of-the-art results in question answering and summarization tasks [BIBREF208].  The Reformer model introduces techniques such as Locality-Sensitive Hashing (LSH) attention and reversible residual layers, which contribute to its memory efficiency and ability to handle long sequences effectively, demonstrating comparable performance to standard Transformers across various NLP tasks [BIBREF209].Performers leverage the FAVOR+ mechanism to achieve nearly linear time complexity and sub-quadratic memory consumption, enhancing both efficiency and accuracy in large-scale applications, including protein sequence modeling and pixel-prediction tasks [BIBREF210]. These advancements collectively indicate a promising trajectory for future research, emphasizing the need for continued exploration of efficient architectures and pre-training strategies that can further enhance the capabilities of Transformer-based models in processing long sequences and complex data structures across diverse domains.
\end{custombox}

\FloatBarrier 

\begin{custombox}[frametitle=Evaluation Example 2]
\textbf{Survey Title:} Recent Advances in Natural Language Processing via Large Pre-Trained Language Models: A Survey

\noindent
\textbf{Survey Section:} Data Generation via PLM:Explaining Models' Decisions

\noindent
\textbf{Evaluation Scores:} ROUGE-1: 0.35, ROUGE-2: 0.05, ROUGE-L: 0.13, BERTScore: 0.63, Ref-F1: 0.35, G-Eval: 4.5, CheckEval: 0.11\\

\noindent
\textbf{Ground Truth (Section Text):}\\
 Despite the impressive performance of deep learning models for various NLP tasks, a remaining challenge to widespread adoption is the lack of explanations for the models' decisions. This hinders the development and debugging process, as well as user trust. This is especially true for application domains such as healthcare, security, and online education. As such, a considerable number of approaches have been proposed for explaining deep learning models' behavior, including model-intrinsic BIBREF220 , BIBREF221 , BIBREF222 and model-agnostic approaches BIBREF223 , BIBREF224 , BIBREF225 . While model-intrinsic explanations expose internal model state (e.g. feature importance or attention scores), in model-agnostic (post-hoc) methods, explanations are generated via the model predictions without inspecting the internal state. Generative models are often applied for post-hoc explanations, aiming to obtain either counterexamples BIBREF226 , BIBREF227 , BIBREF228 or natural language texts BIBREF229 , BIBREF230 , BIBREF231 for explaining purposes. Generating counterexamples can shed light on the decision boundaries of the models (i.e. explaining when a model changes its decision), thus improving intepretability. To this end, the generated counterexamples should be close to the decision boundaries so that small modifications result in changing the model predictions. Traditionally, heuristic rules applied to the original inputs create likely counterexamples BIBREF227 , BIBREF232 , BIBREF233 , BIBREF234 . PLMs have been leveraged to generate more diverse examples for better evaluation BIBREF235 , BIBREF228 , BIBREF236 . In particular, BIBREF228 proposes a method based on GPT-2 to generate counterfactuals that are close to the original sentences and entail specific relationships with the original, facilitating label induction (e.g. negation, insertion, shuffle). Concretely, an input sentence is concatenated with a relation label (e.g. negation) and a template consisting of the special tokens [BLANK] to form the prompt for GPT-2 model. For instance, for the sentence \u201c It is great for kids \" and the relation label \u201c negate \", the following prompt is constructed: \u201c It is great for kids. [negation] It is [BLANK] great for [BLANK]. [SEP] \". Next, the GPT-2 model generates answers for the [BLANK] in the template (e.g. \u201c not [ANSWER] children \u201d, separated by the special token [ANSWER] ). To fine-tune the GPT-2 model, non-parallel datasets (e.g. CommonGen, Natural Questions and SQuAD) are automatically processed to find the relations between pairs of sentences and to construct the templates for each relation based on the obtained pairs. It is worth noting that the sentences generated by GPT-2 might have the same label as the original input sentence. In addition, BIBREF228 show that the generated counterexamples can be helpful to improve the performance of the downstream models, e.g. for natural language inference, duplicate question detection, and sentiment analysis. Other research is informing the task of natural language explanation generation , where the goal is to expose the rationale behind the model decisions in automatically generated natural language text. Any approach must critically require that the generated response is faithful to the model behavior. To this end, BIBREF230 propose to first generate the explanations, and then employ the explanations to obtain the final model predictions. They use natural language inference as the task requiring explanations. Label-specific GPT-2 models are fine-tuned over concatenations of corresponding premises, hypotheses, and human-provided explanations, so that at inference, the model generates an explanation based on premise and hypothesis. Next, the explanations together with the premise and the hypothesis are consumed by an explanation processor model (e.g. RoBERTa) to select the most likely label. This process obtains a more faithful explanation for the label choice, compared to traditional prediction-first approaches BIBREF229 . However, this approach does not provide explanations that reference non-selected labels. To address the question of why other labels are not chosen, BIBREF231 exploit counterexamples, deriving them from original samples with heuristic rules. The original samples and counterexamples are provided to GPT-2 to generate an explanation for the question \u201c Why A not B \u201d.\\

\noindent
\textbf{Generated Content (XSum Output):} \\
The exploration of data generation and model explanation through large pre-trained language models (PLMs) has revealed critical insights into the challenges and advancements in the field. A significant challenge in the interpretability of machine learning models, particularly in natural language processing (NLP), lies in the complexity of the models themselves. Explanations must be comprehensible to users with varying levels of expertise, necessitating a balance between accuracy and interpretability. Techniques such as LIME (Local Interpretable Model-agnostic Explanations) and SHAP (SHapley Additive exPlanations) have emerged as effective methods for generating explanations that enhance user trust in model predictions. LIME has demonstrated high recall rates and usability, enabling users to better assess the trustworthiness of individual predictions, while SHAP offers a theoretically robust framework that aligns closely with human intuition, improving the interpretability of complex models [BIBREF220], [BIBREF221].Trust is a pivotal factor in the deployment of machine learning models, as it influences user reliance on model predictions. Explanations play a crucial role in fostering this trust by elucidating the rationale behind predictions, thereby enabling users to make informed decisions. The experiments conducted with human subjects highlighted that while explanations can enhance understanding, they may also reveal underlying issues with model predictions, leading to a decrease in trust when users discern that predictions are based on untrustworthy reasons [BIBREF220]. Furthermore, the model-agnostic nature of explanation methods like LIME and SHAP allows for their application across various classifiers, facilitating feature engineering and improving model performance [BIBREF220], [BIBREF221].Recent advancements in instancewise feature selection and the use of mutual information have further enhanced model interpretation by providing tailored explanations for specific predictions. This approach contrasts with traditional global feature importance metrics, allowing for a more nuanced understanding of model behavior [BIBREF222]. The introduction of multimodal explanation frameworks, which combine visual and textual justifications, has also shown promise in improving interpretability by leveraging the strengths of different modalities [BIBREF223]. Additionally, the development of datasets such as ACT-X and VQA-X has provided valuable resources for evaluating multimodal explanations in activity recognition and visual question answering tasks, thereby advancing the field of explainable AI [BIBREF223].The integration of counterfactual explanations, particularly through frameworks like GYC and MiCE, has introduced innovative methodologies for generating explanations that clarify model predictions. These approaches focus on generating plausible and diverse counterfactuals that can reveal model behavior and facilitate debugging processes. The ability to produce minimal edits that lead to significant changes in model outputs enhances the interpretability of complex models while maintaining fluency and grammaticality in generated explanations [BIBREF235], [BIBREF236]. Overall, the insights gained from these advancements underscore the importance of developing robust methods for explanation generation, which are essential for fostering trust and understanding in machine learning systems deployed in real-world applications.
\end{custombox}

\end{document}